\documentclass[11pt,a4paper]{article}
\usepackage[hyperref]{emnlp-ijcnlp-2019}
\usepackage{times}
\usepackage{latexsym}
\usepackage{tipa}			%IPA
\usepackage{arydshln}		%dashed line in tabular
\usepackage{tikz}			%drawing
\usepackage{tikz-qtree}		%tree with arrows
\usepackage{bussproofs}		%sequents
\usepackage{stmaryrd}		%semantics brackets (\llbracket, \rrbracket)
\usepackage[linguistics]{forest}	%trees
\forestset{qtree/.style={for tree={parent anchor=south, 
          child anchor=north,align=center,inner sep=0pt}}}
\usepackage{multicol}		%columns
\usepackage{multirow}		%mulit-row cells in tabular
\usepackage{graphics}		%scalebox and images
\usepackage{adjustbox}		%get figures in center
\usepackage{natbib}			%bibliography
\usepackage{linguex}		%examples and glosses
\usepackage{soul}			%strikeout and underline
\usepackage{pbox}			%box, especially for sequents
\usepackage{amsmath}		%math
\usepackage{amssymb}		%for the lozenge symbol
\usepackage{pifont}			%checkmark, \ding{##}
\usepackage{outlines}		%for outlines \1, \2, \3, \4
\usepackage{stackrel}		%stacking math symbols
\usepackage{caption}		%caption table
\usepackage{xcolor}			%text color
\usepackage{hyperref}		%hyperlinks
\usepackage{bm}				%boldface in math mode
\usepackage{hyperref}		%links and such
\usepackage{csquotes}		%in-line quotes
\usepackage{array}			%useful in general, I forget why I imported
\usepackage{booktabs}		%pretty lines in tabular (\toprule, etc)
\usepackage{url}			%makes URLs texttt
\usepackage{subfiles}
\usepackage{lscape}
\usepackage{titlesec}
\usepackage{url}
\usepackage{linguex}
\usepackage{graphicx}
\usepackage{amssymb}% http://ctan.org/pkg/amssymb
\usepackage{pifont}% http://ctan.org/pkg/pifont
\newcommand{\cmark}{\ding{51}}%
\newcommand{\xmark}{\ding{55}}%

\aclfinalcopy % Uncomment this line for the final submission
\title{Linguistic Analysis of Pretrained Sentence Encoders with Acceptability Judgments}

\author{
Alex Warstadt$^{1}$\\
\texttt{\small warstadt@nyu.edu}
\And
Samuel R.~Bowman$^{1,2,3}$\\
\texttt{\small bowman@nyu.edu}
\AND
$^{1}$\normalfont Dept. of Linguistics\\New York University\\10 Washington Place\\New York, NY 10003\And
$^{2}$\normalfont Dept. of Computer Science\\New York University\\60 Fifth Avenue\\New York, NY 10011\And
$^{3}$\normalfont Center for Data Science\\New York University\\60 Fifth Avenue\\New York, NY 10011
} 

\date{}

\begin{document}
\maketitle
\begin{abstract}
% Pretrained sentence encoders like BERT \cite{devlin2018bert} learn to make near-human judgments about the grammatical acceptability of sentences in the Corpus of Linguistic Acceptability \cite[CoLA;][]{warstadt2018neural}. 
Recent work on evaluating grammatical knowledge in pretrained sentence encoders gives a fine-grained view of a small number of phenomena. We introduce a new analysis dataset that also has broad coverage of linguistic phenomena. We annotate the development set of the Corpus of Linguistic Acceptability \cite[CoLA;][]{warstadt2018neural} for the presence of 13 classes of syntactic phenomena including various forms of argument alternations, movement, and modification. We use this analysis set to investigate the grammatical knowledge of three pretrained encoders: BERT \cite{devlin2018bert}, GPT \cite{radford2018improving}, and the BiLSTM baseline from \citeauthor{warstadt2018neural} 
We find that these models have a strong command of complex or non-canonical argument structures like ditransitives (\emph{Sue gave Dan a book}) and passives (\emph{The book was read}). Sentences with long-distance dependencies like questions (\emph{What do you think I ate?}) challenge all models, but for these, BERT and GPT have a distinct advantage over the baseline. We conclude that recent sentence encoders, despite showing near-human performance on acceptability classification overall,  still fail to make fine-grained grammaticality distinctions for many complex syntactic structures.
\end{abstract}

\section{Introduction}

Models for sentence understanding such as BERT \cite{devlin2018bert} and GPT \cite{radford2018improving} are becoming more effective and ubiquitous, leading to a rise in new fine-grained datasets for evaluating their knowledge of grammar. Such evaluations are important for both guiding the development of more robust models and answering theoretical questions about what grammatical concepts are can be acquired through data-driven learning.
To date, most evaluation datasets consist of constructed sentences illustrating a highly specific kind of grammatical contrast \cite{ettinger2016probing,ettinger2018assessing,marvin2018targeted,wilcox2018rnn,wilcox2019structural,futrell2019rnns}, or naturalistic sentences labeled for a particular grammatical feature \cite{linzen2016assessing,shi2016syntax,conneau2017supervised,conneau2018cram}. These evaluation datasets are useful for studying a single phenomenon such as subject-verb agreement in depth \cite{linzen2016assessing}, but fail to test domain general grammatical knowledge. 

By contrast, the Corpus of Linguistic Acceptability\footnote{The original CoLA can be downloaded here: \href{https://nyu-mll.github.io/CoLA/\#}{\url{https://nyu-mll.github.io/CoLA/\#}}} \cite[CoLA;][]{warstadt2018neural} tests these models' ability to judge the grammatical acceptability of sentences in a wide domain. CoLA is a dataset of over 10k example sentences labeled for acceptability and sampled from linguistics publications discussing numerous linguistic phenomena. However, in its original formulation, CoLA does not distinguish between phenomena, making it difficult to analyze a model's knowledge a particular phenomenon.

In this work, we augment CoLA with a new syntactically annotated evaluation set\footnote{The grammatically annotated CoLA can be downloaded here: \href{https://nyu-mll.github.io/CoLA/\#grammatical_annotations}{\url{https://nyu-mll.github.io/CoLA/\#grammatical_annotations}}} that is both fine-grained and domain general. All 1043 examples in the CoLA development set are labeled with expert annotations that indicate the presence or absence of 13 classes of syntactic phenomena, each defined as a union of several more specific phenomena. This resource makes it uniquely easy to conduct analyses whose conclusions can be directly interpreted in terms of modern linguistic theory, since CoLA data is sampled from example sentences in mainstream linguistics publications and our annotations adapt concepts from that literature.

We use our analysis set to assess GPT and BERT, which achieve near-human performance \cite{nangia2019conservative} on many tasks in the GLUE benchmark \cite{wang2018glue}, including CoLA. We treat acceptability classification as a probing task \citep{adi2017fine}, in which a small classifier is trained on CoLA on top of pretrained encoders and tested on the analysis set. We compare these models to a baseline pretrained BiLSTM model released by \newcite{warstadt2018neural} with CoLA.

Our results identify specific syntactic features that make sentences harder to classify, such as long distance dependencies (\emph{What do you think I ate?}), and others that have little effect on difficulty, such as non-canonical argument structures like passives (\emph{The book was read}). Furthermore, some constructions highlight or minimize the differences between models. For example, GPT and BERT far out perform the BiLSTM on movement phenomena such as clefts (\emph{It is Bo that left}), yet have no advantage on sentences with adjuncts (\emph{Sue exercises in the morning}). We wish to exercise caution in interpreting these results since it is not clear to what extent an encoder's failure on a particular phenomenon is due to a weakness of the encoder rather than the training data or probing classifier. However, this caveat applies to varying degrees to all probing resources. In this context of other similar linguistically informed datasets, our analysis set addresses the critical need for a evaluation task with wide coverage of linguistic phenomena.

\section{Related Work}

\paragraph{Sentence Encoders}

\setlength{\tabcolsep}{2.5pt}
\begin{table*}[ht]
\small
    \centering
    \begin{tabular}{l p{0.50\textwidth} c c c c c c c c c c c c c c}
    \toprule
\rotatebox{90}{Acceptability} & \rotatebox{90}{Sentence} & 
\rotatebox{90}{Simple} & 
\rotatebox{90}{Locative} & 
\rotatebox{90}{PP Arg-VP} & 
\rotatebox{90}{High Arity} & 
\rotatebox{90}{Passive} & 
\rotatebox{90}{Binding:Other} & 
\rotatebox{90}{Emb Q} & 
\rotatebox{90}{Complex QP} & 
\rotatebox{90}{Modal} & 
\rotatebox{90}{Raising} & 
\rotatebox{90}{Trans-Adj} & 
\rotatebox{90}{Coord} & 
\rotatebox{90}{Ellipsis/Anaphor} & 
\rotatebox{90}{Comparative} \\\midrule
\cmark & The magazines were sent by Mary to herself. &  &  & \xmark & \xmark & \xmark &  &  &  &  &  &  &  &  &\\
\cmark & John can kick the ball. &  &  &  &  &  &  &  &  & \xmark &  &  &  &  &  \\
* & I know that Meg's attracted to Harry, but they don't know who. &  &  & \xmark &  &  &  & \xmark &  &  &  &  & \xmark & \xmark &  \\
\cmark & They kicked them & \xmark &  &  &  &  & \xmark &  &  &  &  &  &  &  &  \\
\cmark & Which topic did you choose without getting his approval? &  &  &  &  &  &  &  & \xmark &  &  &  &  &  & \\
* & It was believed to be illegal by them to do that. &  &  &  & \xmark & \xmark &  &  &  &  & \xmark & \xmark &  &  & \\
* & Us love they. & \xmark &  &  &  &  &  &  &  &  &  &  &  &  &  \\
* & The more does Bill smoke, the more Susan hates him. &  &  &  &  &  & \xmark &  &  &  &  &  &  &  & \xmark \\
\cmark & I ate a salad that was filled with lima beans. &  &  & \xmark &  & \xmark &  &  &  &  &  &  &  &  &  \\
\cmark & That surprised me. & \xmark &  &  &  &  &  &  &  &  &  &  &  &  &  \\\bottomrule
    \end{tabular}
    \caption{A random sample of sentences from the CoLA development set, shown with their original acceptability labels (\cmark = acceptable, *=unacceptable) and a subset of our new phenomenon-level annotations from the set of finer-grained features.}
    \label{tab:cola sample}
\end{table*}
\setlength{\tabcolsep}{6pt}

Recent research tries to reproduce the success of robust pretrained word embeddings \citep{mikolov2013word2vec,pennington2014glove} at the sentence level, in the form of reusable sentence encoders with pretrained weights. Current state-of-the-art sentence encoders are pretrained on language modeling or related self-supervised tasks. Among these, ELMo \cite{peters2018elmo} uses a BiLSTM architecture, while GPT \cite{radford2018improving} and BERT \cite{devlin2018bert} use the newer attention-based Transformer architecture \cite{vaswani2017attention}. Unlike most earlier approaches where the weights of the encoder are frozen after pretraining, the last two fine-tune the encoder on the downstream task. They are among the top performing\footnote{The current top performing model \cite{liu2019improving} is based on BERT, but includes ensembling and extra fine-tuning.} models on the GLUE benchmark, an evaluation suite for general-purpose sentence understanding models like these, which is built around a set of nine sentence-level classification tasks \cite{wang2018glue}.

\paragraph{Probing Sentence Representations}

The evaluation and analysis of sentence representations is an active area of research. Most of this work to date focuses on in-depth investigation of a particular phenomenon. A popular evaluation technique uses probing tasks in which a small probing classifier is trained to identify particular syntactic and surface features of a sentence based on a sentence representation. 

Some datasets for probing tasks label naturally occurring data with the relevant features. For instance, \citet{shi2016syntax} label sentences with features such as past/present tense and active/passive voice.  \citet{linzen2016assessing} label present tense verbs for whether they have singular or plural agreement marking. \citet{adi2017fine} label sentences for length and word content. \citet{conneau2018cram} label sentences for syntactic depth and morphological number. 

Another common method is to semi-automatically generate data manipulating a small set of grammatical features. For instance, \citet{ettinger2016probing,ettinger2018assessing} build datasets of this kind to test whether sentence encoders represent the scope of negation and semantic roles, and \citet{kann2019verb} do so to test whether word and sentence encoders representations information about verbal argument structure.

\paragraph{CoLA \& Acceptability Classification}

The Corpus of Linguistic Acceptability \cite{warstadt2018neural} is a dataset of 10k example sentences including expert annotations for grammatical acceptability. The sentences are examples taken from 23 theoretical linguistics publications and represent a wide array of phenomena discussed in that literature. Such example sentences are usually labeled for acceptability by their authors or a small group of native English speakers. A small random sample of the CoLA development set (with our added annotations) can be seen in Table \ref{tab:cola sample}.

Within computational linguistics, the acceptability classification task has been explored previously:
\newcite{lawrence2000grammatical} train RNNs to do acceptability classification over sequences of POS tags corresponding to example sentences from a syntax textbook. \newcite{wagner2009grammaticality} also train RNNs, but using naturally occurring sentences that have been automatically manipulated to be unacceptable. \newcite{lau2016cognitive} predict acceptability from language model probabilities, applying this technique to sentences from a syntax textbook, and sentences which were translated round-trip through various languages.

\citeauthor{lau2016cognitive}\ also attempt to model gradient crowd-sourced acceptability judgments, reflecting an ongoing debate about whether binary expert judgments like those in CoLA are reliable \cite{gibson2010quantitative,sprouse2012adger}. We remain agnostic as to the role of binary judgments in linguistic theory, but note that \newcite{warstadt2018neural} and \newcite{nangia2019conservative} measure expert and non-expert human performance, respectively, on subsets of CoLA (see Table \ref{tab:results} for the former's results), both finding that new human annotators, while not in perfect agreement with the judgments in CoLA, still agree well and outperform the best neural network models.

\section{Analysis Set}

We introduce a grammatically annotated version of the entire CoLA development set to facilitate detailed error analysis of acceptability classifiers. These 1043 sentences are labeled with 13 major features, further divided into 59 minor features. Each feature marks the presence of a particular phenomenon or class of phenomena in the sentence. Each minor feature belongs to
a single major feature. A sentence belongs to a
major feature if it belongs to one or more of the
relevant minor features. The supplementary materials include descriptions of each feature along
with examples and the criteria used for annotation.
%Sentences with multiple relevant minor features are counted only once for each major feature in our evaluations. 

% The intended purpose of this resource is to make generalizations about what syntactic features of sentences make them easier or harder for acceptability classifiers to judge, and to develop better intuitions about how reliably sentence embeddings encode these features. 

% The sentences in the CoLA development set are split nearly equally between in-domain and out-of-domain sentences, and each is similarly imbalanced with about $69\%$ of examples acceptable. 
The major and minor features are listed in Table \ref{tab:categories}, and are fully documented in the Appendix. The average sentence is positively labeled with 3.22 major features (SD=1.66) on average, and the average a major feature is present in 224 sentences (SD=112). 
The average sentence is positively labeled with 4.31 minor features (SD=2.59). The average minor feature is present in 71.3 sentences (SD=54.7). Every sentence is labeled with at least one feature. Sentences without any obvious phenomena of interest are labeled \textsc{simple}.

\begin{table*}[t]
    \centering
    \small
    \begin{tabular}{p{0.17\textwidth} p{0.75\textwidth}}
    \toprule
    \textbf{Major Feature ($n$)} & \textbf{Minor Features ($n$)} \\\midrule
    \textbf{Simple (87)} & Simple (87) \\
    \textbf{Pred (256)} & Copula (187),	Pred/SC (45),	Result/Depictive (26)  \\
    \textbf{Adjunct (226)}	&VP Adjunct (162), Misc Adjunct (75), Locative (69), NP Adjunct (52), Temporal (49), Particle (33)\\
    \textbf{Arg Types (428)} & PP Arg VP (242), Oblique (141), PP Arg NP/AP (81),	Expletive (78),	by-Phrase (58)\\
    \textbf{Arg Altern (421)} & High Arity (253),	Passive (114), Drop Arg (112),	Add Arg (91)\\
    \textbf{Bind (121)} & Binding:Other (62), Binding:Refl (60) \\
    \textbf{Question (222)} & Emb Q (99),  Pied Piping (80), Rel Clause (76), Matrix Q (56), Island (22) \\
    \textbf{Comp Clause	(190)} & CP Arg VP (110),  No C-izer (41), Deep Embed (30), CP Arg NP/AP (26), Non-finite CP (24), CP Subj (15)\\
    \textbf{Auxiliary (340)}	& Aux (201), Modal (134), Neg (111),   Psuedo-Aux (26)\\
    \textbf{to-VP (170)	}& Control (80), Non-finite VP Misc (38), VP Arg NP/AP	(33), VP+Extract (26), Raising (19) \\
  \textbf{N, Adj (278) }& Compx NP (106), Rel NP (65), Deverbal (53), Trans Adj (39),  NNCompd (35), Rel Adj (26),  Trans NP (21)\\
    \textbf{S-Syntax (286)}& Coord (158), Ellipsis/Anaphor (118), Dislocation (56),  Subordinate/Cond (41), Info Struc (31), S-Adjunct (30), Frag/Paren (9)\\
  \textbf{Determiner	(178)} & Quantifier (139), NPI/FCI (29), Comparative (25), Partitive (18)\\\bottomrule
    \end{tabular}
    \caption{Major features and their associated minor features (with number of occurrences $n$).}
    \label{tab:categories}
\end{table*}

\subsection{Annotation}

The sentences were annotated manually by one of the authors, who is trained in formal linguistics and linguistic annotation. The features were developed in a trial stage, in which the annotator performed a similar annotation with different annotation schema for several hundred sentences from CoLA not belonging to the development set. 

\subsection{Feature Descriptions}

Here we briefly summarize the feature set in order of the major features. These constructions are well-studied in syntax, and further background can be found in textbooks such as \newcite{adger2003core} and \newcite{sportiche2013introduction}.

\paragraph{Simple} This major feature contains only one minor feature, \textsc{simple}, including sentences where the subject and predicate are unmodified (\emph{Bo ate an apple}.)

\paragraph{Pred(icate)} These three features correspond to predicative phrases, including copular constructions (\emph{Bo is awake.}), small clauses (\emph{I saw \underline{Bo jump}}), and resultatives/depictives (\emph{Bo wiped \underline{the table clean}}).

\paragraph{Adjunct} These six features mark various kinds of optional modifiers, including modifiers of NPs (\emph{The boy \underline{with blue eyes} gasped}) or VPs (\emph{Bo sang \underline{with Jo}}), and temporal (\emph{Bo awoke \underline{at dawn}}) or locative (\emph{Bo jumped \underline{on the bed}}) adjuncts.

\paragraph{Argument types} These five features identify syntactically selected arguments, differentiating, for example, obliques (\emph{I gave a book \underline{to Bo}}), PP arguments of NPs and VPs (\emph{Bo voted \underline{for Jones}}), and expletives (\emph{\underline{It} seems that Bo left}).

\paragraph{Argument Alternations} These four features mark VPs with unusual argument structures, including added arguments (\emph{I baked \underline{Bo} a cake}) or dropped arguments (\emph{Bo \underline{knows}}), and the passive (\emph{I was \underline{applauded}}).

\paragraph{Bind} These are two minor features, one for bound reflexives (\emph{Bo loves \underline{himself}}), and one for other bound pronouns (\emph{Bo thinks \underline{he} won}).

\paragraph{Question} These five features apply to sentences with question-like properties. They mark whether the interrogative is an embedded clause (\emph{I know \underline{who you are}}), a matrix clause (\emph{Who are you?}), or a relative clause (\emph{Bo saw the guy \underline{who left}}); whether it contains an island out of which extraction is unacceptable (\emph{*What was \underline{a picture of} hanging on the wall?})\footnote{Following standard notation in linguistics, the `*' precedes sentences that are not grammatically acceptable.}; or whether there is pied-piping or a multi-word \emph{wh}-expressions (\emph{\underline{With whom} did you eat?}).

\paragraph{Comp(lement) Clause} These six features apply to various complement clauses (CPs), including subject CPs (\emph{\underline{That Bo won} is odd}); CP arguments of VPs or NPs/APs (\emph{The fact \underline{that Bo won}}); CPs missing a complementizer (\emph{I think \underline{Bo's crazy}}); or non-finite CPs (\emph{This is ready \underline{for you to eat}}).

\paragraph{Aux(iliary)} These four minor features mark the presence of auxiliary or modal verbs (\emph{I \underline{can} win}), negation, or ``pseudo-auxiliaries'' (\emph{I \underline{have to} win}).

\paragraph{to-VP} These five features mark various infinitival embedded VPs, including control VPs (\emph{Bo wants \underline{to win}}); raising VPs (\emph{Bo seemed \underline{to fly}}); VP arguments of NPs or APs (\emph{Bo is eager \underline{to eat}}); and VPs with extraction (e.g. \emph{This is easy \underline{to read} \underline{\phantom{ts}} }).

\paragraph{N(oun), Adj(ective)} These seven features mark complex NPs and APs, including ones with PP arguments (\emph{Bo is \underline{fond of Mo}}), or CP/VP arguments; noun-noun compounds (\emph{Bo ate \underline{mud pie}}); modified NPs, and NPs derived from verbs (\emph{\underline{Baking} is fun}).

\paragraph{S-Syntax} These seven features mark various unrelated syntactic constructions, including dislocated phrases (\emph{The boy left \underline{who was here earlier}}); movement related to focus or information structure (\emph{\underline{This} I've gotta see \underline{\phantom{this}}}); coordination, subordinate clauses, and ellipsis (\emph{I can't}); or sentence-level adjuncts (\emph{\underline{Apparently}, it's raining}).

\paragraph{Determiner} These four features mark various determiners, including quantifiers, partitives (\emph{two of the boys}), negative polarity items (\emph{I *do/don't have \underline{any} pie}), and comparative constructions.

\subsection{Correlations}

These features are overlapping and in many cases are correlated, so not all results from using this analysis set will be independent. We analyzed the between-feature pairwise Matthews Correlation Coefficient \cite[MCC;][]{matthews1975correlation} of the 63 minor features (giving 1953 pairs), and of the 15 major features (giving 105 pairs). MCC is a special case of Pearson's $r$ for Boolean variables.\footnote{MCC measures correlation of two binary distributions, giving a value between -1 and 1. On average, any two unrelated distributions will have a score of 0, regardless of class imbalance. This is contrast to metrics like accuracy or F1, which favor classifiers with a majority-class bias.} These results are summarized in Table \ref{tab:corr_cats}. Regarding the minor features, 60 pairs had a correlation of 0.2 or greater and 15 had a correlation of 0.5 or greater. Turning to the major features, 6 pairs had a correlation of 0.2 or greater, and 2 had an anti-correlation of greater magnitude than -0.2.

We can see at least three reasons for these observed correlations. First, some features have overlapping definitions; for example \textsc{expletive} is a strict subset of \textsc{add arg} because expletive arguments (e.g. \emph{There are birds singing}) are by definition non-canonical. Similarly, the strong anti-correlation between \textsc{simple} and the two features related to argument structure, \textsc{argument types} and \textsc{arg altern}, follows from the definition of \textsc{simple}, which explicitly excludes sentences with unusual argument structure. Second, grammatical facts of English drive the correlation between, for instance, \textsc{question} and \textsc{aux}, because main-clause questions in English require subject-aux inversion. Third, the unusually high correlation of, for example, \textsc{Emb-Q} and \textsc{ellipsis/anaphor}, can be attributed largely to a bias in a particular source in CoLA, \citet{chung1995sluicing}, which is an article about the sluicing construction involving ellipsis of an embedded interrogative (e.g. \emph{I saw someone, but I don't know who}). 
This third case highlights a limitation of this analysis set. The set of examples associated with a particular feature is not a controlled set designed to test knowledge of that particular construction, but rather a sample of sentences from the linguistics literature, and as such may not be full representative of the construction in question. However, this cost comes with the advantage that results on the analysis set can be directly connected to relevant linguistics literature.

% Finally, two strongest anti-correlations between major features are between \textsc{simple} and the two features related to argument structure, \textsc{argument types} and \textsc{arg altern}. This follows from the definition of \textsc{simple}, which excludes any sentence containing a large number or unusual configuration of arguments.

\begin{table}[t]
\small
    \centering
    \begin{tabular}{l l r}
\toprule
\textbf{Label 1} & \textbf{Label 2} & \textbf{MCC}\\\midrule
\multicolumn{3}{c}{\textbf{Major Features}}\\\midrule
Argument Types &	Arg Altern &	0.406\\
Question &	Auxiliary &	0.273\\
Question &	S-Syntax &	0.232\\
Predicate &	N, Adj &	0.231\\
Auxiliary &	S-Syntax &	0.224\\
Question &	N, Adj &	0.211\\
Simple &	Arg Altern &	-0.227\\
Simple &	Argument Types &	-0.238\\\midrule
\multicolumn{3}{c}{\textbf{Minor Features}}\\\midrule
PP Arg NP/AP &	Rel NP &	0.755\\
by-Phrase &	Passive &	0.679\\
Coord &	Ellipsis/Anaphor &	0.634\\
VP Arg NP/AP &	Trans Adj &	0.628\\
NP Adjunct &	Compx NP &	0.623\\
Oblique &	High Arity &	0.620\\
RC &	Compx NP &	0.565\\
Expletive &	Add Arg &	0.558\\
CP Arg NP/AP &	Trans NP &	0.546\\
PP Arg NP/AP	& Rel Adj &	0.528\\
% VP Adjunct &	Temporal &	0.518\\
% Oblique	& PP Arg VP &	0.507\\
% VP Adjunct &	Misc Adjunct &	0.485\\
% Emb Q &	Ellipsis/Anaphor &	0.463\\
% VP Adjunct &	Locative &	0.418\\
% Drop Arg &	Passive &	0.414\\
% Matrix Q &	Pied Piping &	0.411\\
\bottomrule
    \end{tabular}
    \caption{Correlation (MCC) of features in the annotated analysis set. We display only the correlations with the greatest magnitude.}
    \label{tab:corr_cats}
\end{table}

\begin{figure*}[h]
    \centering
    \includegraphics[width=1\textwidth]{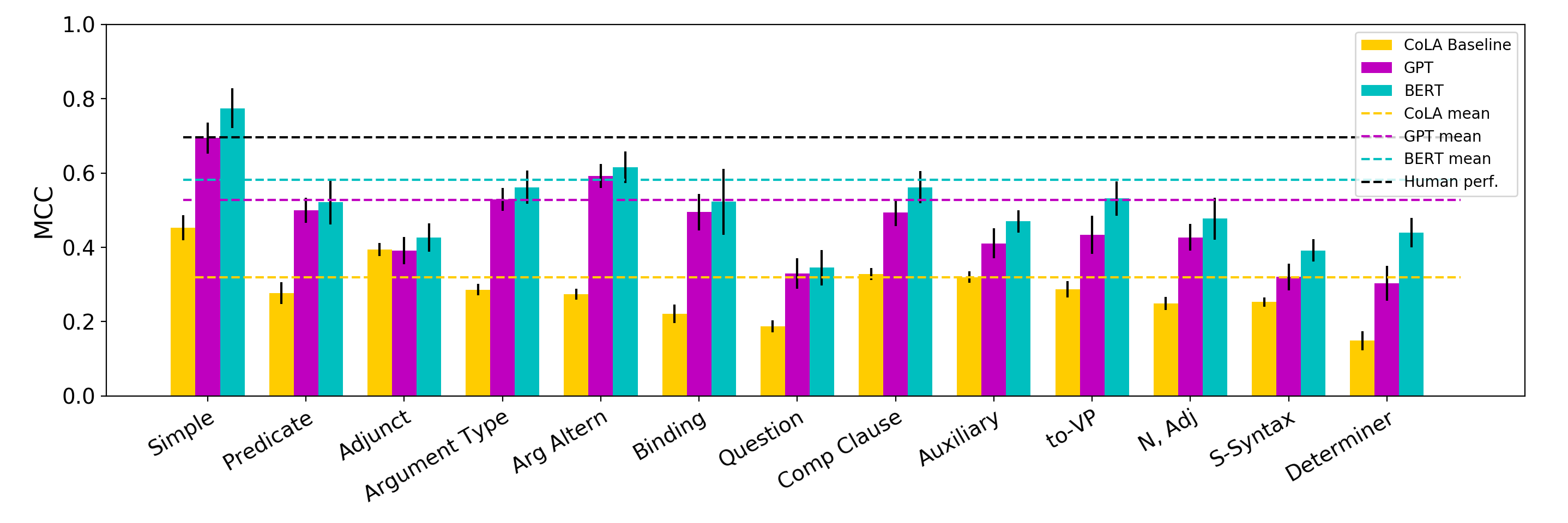}
    \caption{Performance (MCC) on our analysis set by major feature. Dashed lines show mean performance on the entire CoLA development set. Error bars mark the mean $\pm 1$ standard deviation. From left to right, performance for each feature is given for the CoLA Baseline, OpenAI GPT, and BERT.}
    \label{fig:major}
\end{figure*}

\section{Models Evaluated}

We train MLP acceptability classifiers for CoLA on top of three sentence encoders: (1) CoLA's pretrained BiLSTM baseline encoder, (2) OpenAI GPT, and (3) BERT. We use publicly available pretrained sentence encoders.\footnote{\noindent CoLA baseline: \url{https://github.com/nyu-mll/CoLA-baselines}\\
OpenAI GPT: \url{https://github.com/openai/finetune-transformer-lm}\\
BERT: \url{https://github.com/google-research/bert}
}

\paragraph{LSTM Encoder: CoLA Baseline}
The CoLA baseline model is the sentence encoder with the highest performance on CoLA from \citeauthor{warstadt2018neural} The encoder uses a BiLSTM, which reads the sentence word-by-word in both directions, with max-pooling over the hidden states. Similar to ELMo \cite{peters2018elmo}, the inputs to the BiLSTM are the hidden states of a language model (only a forward language model is used in contrast with ELMo). The encoder is trained on a real/fake discrimination task which requires it to identify whether a sentence is naturally occurring or automatically generated. We train acceptability classifiers on CoLA using the CoLA baselines codebase with 20 random restarts, following the original authors' transfer-learning approach: The sentence encoder's weights are frozen, and the sentence embedding serves as input to an MLP with a single hidden layer. All hyperparameters are held constant across restarts.
% the sentence encoder's weights are frozen, and the sentence embedding ($d=1056$) serves as input to an MLP with a single hidden layer ($d=256$).

\paragraph{Transformer Encoders: GPT and BERT} In contrast with recurrent models, GPT and BERT use a self attention mechanism which combines representations for each (possibly non-adjacent) pair of words to give a sentence embedding. GPT is trained using a standard language modeling task, while BERT is trained with masked language modeling and next sentence prediction tasks. We use BERT$_\textsc{large}$. For each encoder, we train 20 random restarts on CoLA feeding the pretrained models published by these authors into a single output layer, using code which will be released upon publication. Following the methods of the original authors, we fine-tune the encoders during training on CoLA. All hyperparameters are held constant across restarts.

\section{Results}

\subsection{Overall CoLA Results}

The overall performance of the three sentence encoders is shown in Table \ref{tab:results}. Following \citeauthor{warstadt2018neural}, performance on CoLA is measured using MCC. We present the best single restart for each encoder, the mean over restarts for an encoder, and the result of ensembling the restarts for a given encoder, i.e. the majority classification for a given sentence, or \textit{acceptable} if tied.\footnote{Because we use the development set for analysis, we do not use it to weight models for weighted ensembling.} For BERT results, we exclude 5 out of the 20 restarts because they were degenerate (MCC=0).

Across the board, BERT outperforms GPT, which outperforms the CoLA baseline. However, BERT and GPT are much closer in performance than they are to CoLA baseline. While ensemble performance exceeded the average for BERT and GPT, it did not outperform the best single model. 
%This indicates

\begin{table}[t]
\small
    \centering
    \begin{tabular}{l r r r}
\toprule
& \textbf{Mean (STD)} & \textbf{Max} & \textbf{Ensemble}\\\midrule
CoLA & 0.320 (0.007) & 0.330 & 0.320\\
GPT & 0.528 (0.023) & 0.575 & 0.567\\
BERT & \textbf{0.582} (0.032) & \textbf{0.622} & \textbf{0.601}\\\midrule
Human & \emph{0.697 (0.042)} & \emph{0.726} & \emph{0.761}\\
\bottomrule
    \end{tabular}
    \caption{Performance (MCC) on the CoLA test set, including mean over restarts of a given model with standard deviation, max over restarts, and majority prediction over restarts. Human agreement is measured by \citeauthor{warstadt2018neural}.\footnote{Human annotations are available at \href{https://nyu-mll.github.io/CoLA/}{\url{https://nyu-mll.github.io/CoLA/}}}}
    \label{tab:results}
\end{table}

\subsection{Analysis Set Results}

\begin{figure*}[h]
    \centering
    \includegraphics[width=\textwidth]{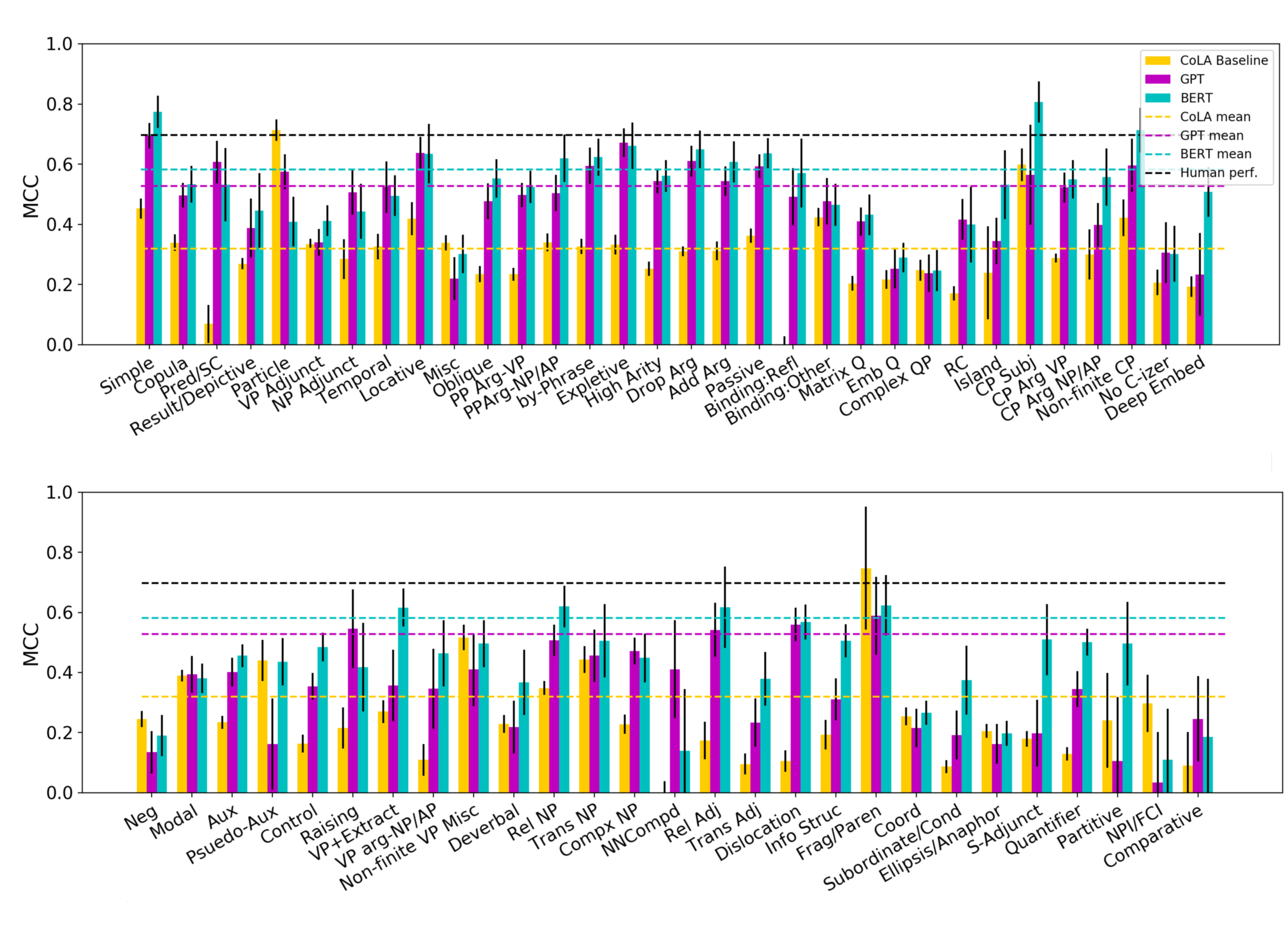}
    \caption{Performance (MCC) on our analysis set by minor feature. Dashed lines show mean performance on the entire CoLA development set. Error bars mark the mean $\pm 1$ standard deviation. From left to right, performance for each feature is given for the CoLA Baseline, OpenAI GPT, and BERT.}
    \label{fig:minor}
\end{figure*}

The results for the major and minor features are shown in Figures \ref{fig:major} and \ref{fig:minor}, respectively. For each feature, we measure the MCC of the sentences including that feature. We plot the mean of these results across the different restarts for each model.

Comparison across features reveals that the presence of certain features has a large effect on performance, and we comment on some patterns below. Within a given feature, the effect of model type is overwhelmingly stable, and resembles the overall difference in performance. However, we observe several interactions, i.e. specific features where the relative performance of models does not track their overall relative performance. 
In interpreting these results, we caution against drawing strong conclusions from rare minor features. For this reason, we do not discuss any results for features appearing in fewer than 50 sentences. Furthermore, we cannot conclude with certainty that any particular mode of success or failure reflects what the information in the pretrained encoder, rather than what sorts of contrasts are easy or hard to learn from the CoLA training data. However, we consider results for major features more likely to be reliable due to the large number and variety of sentences with each label.

\paragraph{Comparing Features}

Among the major features (Figure \ref{fig:major}), performance is universally highest on the \textsc{simple} sentences, and is higher than each model's overall performance. Otherwise we find that a model's performance on sentences of a given feature is on par with or lower than its overall performance, reflecting the fact that features mark the presence of unusual or complex syntactic structure. Performance is also high (and close to overall performance) on sentences with marked argument structures (\textsc{Argument Types} and \textsc{Arg(ument) Alt(ernation)}), indicating that argument structure is relatively easy to learn.

Comparing different kinds of embedded content, we observe higher performance on sentences with embedded clauses (major feature=\textsc{Comp Clause}) embedded VPs (major feature=\textsc{to-VP}) than on sentences with embedded interrogatives (minor features=\textsc{Emb-Q}, \textsc{Rel Clause}). Interrogatives are quite challenging in general (major feature=\textsc{Question}). Sentences with question-like syntax may be difficult because they usually involve extraction of a \emph{wh}-word, creating a long-distance dependency between the \emph{wh}-word and its extraction site, which may be difficult for models to recognize.

% However, in other cases this is intuitively less likely. Despite accounting for about 10\% of examples each, performance on \textsc{emb Q} is about half that on \textsc{passive}, despite the fact that a large number of passive sentences in the analysis set come from a research paper on the passive illustrating many subtle facts about that construction. 

\paragraph{Comparing Models}

Comparing within-feature performance of the three encoders to their overall performance, we find they have differing strengths and weaknesses. BERT and GPT generally far outperform the CoLA baseline, with BERT performing best in most cases.
%BERT stands outover other models\textsc{Deep Embed}, which includes challenging sentences with doubly-embedded clauses, as well as in several features involving extraction (i.e. long-distance dependencies) such as \textsc{VP+Extract} and \textsc{Info-Struc}. 
BERT and GPT have a particularly large advantage in sentences involving long-distance dependencies. They outperform the CoLA baseline by an especially wide margin on \textsc{Bind:Refl}, which involves establishing a dependency between a reflexive and its antecedent (\emph{\underline{Bo} tries to love \underline{himself}}), as well as \textsc{dislocation}, in which expressions are separated from their dependents (\emph{Bo \underline{practiced} on the train \underline{an important presentation}}). The advantage of BERT and GPT may be due in part to their use of the Transformer architecture. Unlike the BiLSTM used by the CoLA baseline, the Transformer uses a self-attention mechanism that associates all pairs of words regardless of distance.

In some specific instances, we do not observe the usual pattern of BERT outperforming GPT and both far outperforming the CoLA baseline, revealing possible idiosyncrasies of the sentence representations they output. For instance, the CoLA baseline performs on par with the others on the major feature \textsc{adjunct}, especially considering the minor feature \textsc{Particle} (\emph{Bo \underline{looked} the \underline{word} up}).

\subsection{Length Analysis}

\begin{figure}[t]
    \centering
    \includegraphics[width=0.5\textwidth]{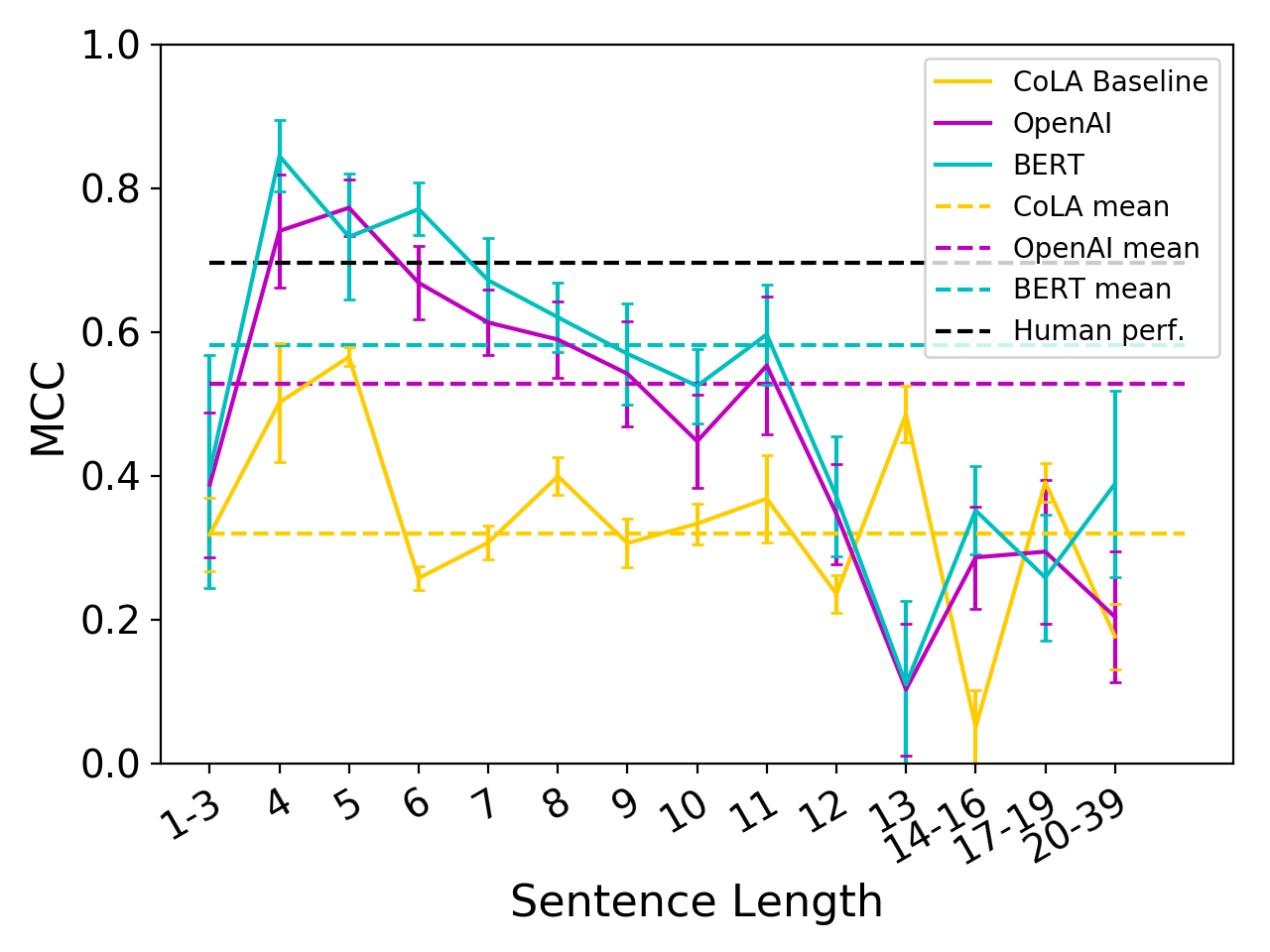}
    \caption{Performance (MCC) on the CoLA analysis set by sentence length.}
    \label{fig:length}
\end{figure}

For comparison, we analyze the effect of sentence length on acceptability classifier performance. The results are shown in Figure \ref{fig:length}. The results for the CoLA baseline are inconsistent, but do drop off as sentence length increases. For BERT and GPT, performance decreases very steadily with length. Exceptions are extremely short sentences (length 1-3), which may be challenging due to insufficient information; and extremely long sentences, where we see a small (but somewhat unreliable) boost in BERT's performance. BERT and GPT are generally quite close in performance, except on the longest sentences, where BERT's performance is considerably better.

\section{Conclusion}

Using a new grammatically annotated analysis set, we identify several syntactic phenomena that are predictive of good or bad performance of current state-of-the-art sentence encoders on CoLA. We also use these results to develop hypotheses about why BERT is successful, and why Transformer models outperform sequence models.

Our findings can guide future work on sentence representation. Transformer models appear to have an advantage over sequence models with long-distance dependencies, but still struggle with these constructions relative to more local phenomena. It stands to reason that this performance gap might be widened by training larger or deeper Transformer models, or training on longer or more complex sentences. This analysis set can be used by engineers interested in evaluating the syntactic knowledge of their encoders.

% that the syntactic phenomena in a sentence are predictive of how well an acceptability classifier is likely to perform on CoLA. Many of these findings are stable when comparing across different model types, and different restarts within the same model type. For example, marked argument structures like the passive do not decrease performance, while the presence of questions, negation, or ellipsis does. We also find various forms of evidence that one of BERT's biggest advantages over other models is in long distance syntax. In addition to outperforming GPT by a wide in several features involving extraction, it also has the greatest advantage over GPT in sentences of 20 words or longer. 

Finally, these findings suggest possible controlled experiments that could confirm whether there is a causal relation between the presence of the syntactic features we single out as interesting and model performance. Our results are purely correlational, and do not mark whether a particular construction is crucial for the acceptability of the sentence. Future experiments following \newcite{ettinger2018assessing} and \newcite{kann2019verb} can semi-automatically generate datasets by manipulating, for example, length of long-distance dependencies, inflectional violations, or the presence of interrogatives, while controlling for factors like sentence length and word choice, in order to determine the extent to which these features impact the quality of sentence representations.

\section*{Acknowledgments}

This material is based upon work supported by the National Science Foundation under Grant No. 1850208. Any opinions, findings, and conclusions or recommendations expressed in this material are those of the author(s) and do not necessarily reflect the views of the National Science Foundation. This project has also benefited from financial support to SB by Samsung Research under the project \textit{Improving Deep Learning using Latent Structure}
and from the donation of a Titan V GPU by NVIDIA Corporation.

\bibliographystyle{acl_natbib}
\bibliography{emnlp-ijcnlp-2019}

\begin{thebibliography}{39}
\expandafter\ifx\csname natexlab\endcsname\relax\def\natexlab#1{#1}\fi

\bibitem[{Adger(2003)}]{adger2003core}
David Adger. 2003.
\newblock \emph{Core Syntax: A Minimalist Approach}.
\newblock Oxford University Press Oxford.

\bibitem[{Adi et~al.(2017)Adi, Kermany, Belinkov, Lavi, and
  Goldberg}]{adi2017fine}
Yossi Adi, Einat Kermany, Yonatan Belinkov, Ofer Lavi, and Yoav Goldberg. 2017.
\newblock Fine-grained analysis of sentence embeddings using auxiliary
  prediction tasks.
\newblock In \emph{Proceedings of ICLR Conference Track. Toulon, France.}

\bibitem[{Carnie(2013)}]{carnie2013syntax}
Andrew Carnie. 2013.
\newblock \emph{Syntax: A Generative Introduction}.
\newblock John Wiley \&amp; Sons.

\bibitem[{Chung et~al.(1995)Chung, Ladusaw, and McCloskey}]{chung1995sluicing}
Sandra Chung, William~A Ladusaw, and James McCloskey. 1995.
\newblock Sluicing and logical form.
\newblock \emph{Natural Language Semantics}, 3(3):239--282.

\bibitem[{Collins(2005)}]{collins2005smuggling}
Chris Collins. 2005.
\newblock A smuggling approach to the passive in {E}nglish.
\newblock \emph{Syntax}, 8(2):81--120.

\bibitem[{Conneau et~al.(2017)Conneau, Kiela, Schwenk, Barrault, and
  Bordes}]{conneau2017supervised}
Alexis Conneau, Douwe Kiela, Holger Schwenk, Lo{\"\i}c Barrault, and Antoine
  Bordes. 2017.
\newblock Supervised learning of universal sentence representations from
  natural language inference data.
\newblock In \emph{Proceedings of the 2017 Conference on Empirical Methods in
  Natural Language Processing}, pages 670--680.

\bibitem[{Conneau et~al.(2018)Conneau, Kruszewski, Lample, Barrault, and
  Baroni}]{conneau2018cram}
Alexis Conneau, German Kruszewski, Guillaume Lample, Lo{\"\i}c Barrault, and
  Marco Baroni. 2018.
\newblock What you can cram into a single \$\&!\#* vector : Probing sentence
  embeddings for linguistic properties.
\newblock \emph{arXiv preprint arXiv:1805.01070}.

\bibitem[{Culicover and Jackendoff(1999)}]{culicover1999comparative}
Peter~W Culicover and Ray Jackendoff. 1999.
\newblock The view from the periphery: The {E}nglish comparative correlative.
\newblock \emph{Linguistic Inquiry}, 30(4):543--571.

\bibitem[{Devlin et~al.(2018)Devlin, Chang, Lee, and
  Toutanova}]{devlin2018bert}
Jacob Devlin, Ming-Wei Chang, Kenton Lee, and Kristina Toutanova. 2018.
\newblock Bert: Pre-training of deep bidirectional transformers for language
  understanding.
\newblock \emph{arXiv preprint arXiv:1810.04805}.

\bibitem[{Ettinger et~al.(2018)Ettinger, Elgohary, Phillips, and
  Resnik}]{ettinger2018assessing}
Allyson Ettinger, Ahmed Elgohary, Colin Phillips, and Philip Resnik. 2018.
\newblock Assessing composition in sentence vector representations.
\newblock In \emph{Proceedings of the 27th International Conference on
  Computational Linguistics}, pages 1790--1801. Association for Computational
  Linguistics.

\bibitem[{Ettinger et~al.(2016)Ettinger, Elgohary, and
  Resnik}]{ettinger2016probing}
Allyson Ettinger, Ahmed Elgohary, and Philip Resnik. 2016.
\newblock Probing for semantic evidence of composition by means of simple
  classification tasks.
\newblock In \emph{Proceedings of the 1st Workshop on Evaluating Vector-Space
  Representations for NLP}, pages 134--139.

\bibitem[{Futrell and Levy(2019)}]{futrell2019rnns}
Richard Futrell and Roger~P Levy. 2019.
\newblock Do rnns learn human-like abstract word order preferences?
\newblock \emph{Proceedings of the Society for Computation in Linguistics},
  2(1):50--59.

\bibitem[{Gibson and Fedorenko(2010)}]{gibson2010quantitative}
Edward Gibson and Evelina Fedorenko. 2010.
\newblock Weak quantitative standards in linguistics research.
\newblock \emph{Trends in Cognitive Sciences}, 14(6):233--234.

\bibitem[{Goldberg and Jackendoff(2004)}]{goldberg2004resultative}
Adele~E Goldberg and Ray Jackendoff. 2004.
\newblock The {E}nglish resultative as a family of constructions.
\newblock \emph{Language}, 80(3):532--568.

\bibitem[{Kadmon and Landman(1993)}]{kadmon1993any}
Nirit Kadmon and Fred Landman. 1993.
\newblock \href {https://doi.org/10.1007/bf00985272} {Any}.
\newblock \emph{Linguistics and Philosophy}, 16(4):353--422.

\bibitem[{Kann et~al.(2019)Kann, Warstadt, Williams, and Bowman}]{kann2019verb}
Katharina Kann, Alex Warstadt, Adina Williams, and Samuel~R Bowman. 2019.
\newblock Verb argument structure alternations in word and sentence embeddings.
\newblock \emph{Proceedings of the Society for Computation in Linguistics},
  2(1):287--297.

\bibitem[{Kim and Sells(2008)}]{kim2008syntax}
Jong-Bok Kim and Peter Sells. 2008.
\newblock \emph{English Syntax: An Introduction}.
\newblock CSLI Publications.

\bibitem[{Lau et~al.(2016)Lau, Clark, and Lappin}]{lau2016cognitive}
Jey~Han Lau, Alexander Clark, and Shalom Lappin. 2016.
\newblock Grammaticality, acceptability, and probability: a probabilistic view
  of linguistic knowledge.
\newblock \emph{Cognitive Science}, 41(5):1202--1241.

\bibitem[{Lawrence et~al.(2000)Lawrence, Giles, and
  Fong}]{lawrence2000grammatical}
Steve Lawrence, C~Lee Giles, and Sandiway Fong. 2000.
\newblock Natural language grammatical inference with recurrent neural
  networks.
\newblock \emph{IEEE Transactions on Knowledge and Data Engineering},
  12(1):126--140.

\bibitem[{Linzen et~al.(2016)Linzen, Dupoux, and
  Goldberg}]{linzen2016assessing}
Tal Linzen, Emmanuel Dupoux, and Yoav Goldberg. 2016.
\newblock Assessing the ability of lstms to learn syntax-sensitive
  dependencies.
\newblock \emph{Transactions of the Association for Computational Linguistics},
  4:521--535.

\bibitem[{Liu et~al.(2019)Liu, He, Chen, and Gao}]{liu2019improving}
Xiaodong Liu, Pengcheng He, Weizhu Chen, and Jianfeng Gao. 2019.
\newblock \href {http://arxiv.org/abs/1904.09482} {Improving multi-task deep
  neural networks via knowledge distillation for natural language
  understanding}.
\newblock \emph{CoRR}, abs/1904.09482.

\bibitem[{Marvin and Linzen(2018)}]{marvin2018targeted}
Rebecca Marvin and Tal Linzen. 2018.
\newblock Targeted syntactic evaluation of language models.
\newblock In \emph{Proceedings of the 2018 Conference on Empirical Methods in
  Natural Language Processing}, pages 1192--1202.

\bibitem[{Matthews(1975)}]{matthews1975correlation}
Brian~W Matthews. 1975.
\newblock Comparison of the predicted and observed secondary structure of t4
  phage lysozyme.
\newblock \emph{Biochimica et Biophysica Acta (BBA)-Protein Structure},
  405(2):442--451.

\bibitem[{Mikolov et~al.(2013)Mikolov, Sutskever, Chen, Corrado, and
  Dean}]{mikolov2013word2vec}
Tomas Mikolov, Ilya Sutskever, Kai Chen, Greg~S Corrado, and Jeff Dean. 2013.
\newblock Distributed representations of words and phrases and their
  compositionality.
\newblock In \emph{Advances in neural information processing systems}, pages
  3111--3119.

\bibitem[{Nangia and Bowman(2019)}]{nangia2019conservative}
Nikita Nangia and Samuel~R Bowman. 2019.
\newblock A conservative human baseline estimate for glue: People still
  (mostly) beat machines.

\bibitem[{Pennington et~al.(2014)Pennington, Socher, and
  Manning}]{pennington2014glove}
Jeffrey Pennington, Richard Socher, and Christopher Manning. 2014.
\newblock {GloVe}: Global vectors for word representation.
\newblock In \emph{Proceedings of the 2014 Conference on Empirical Methods in
  Natural Language Processing (EMNLP)}, pages 1532--1543.

\bibitem[{Peters et~al.(2018)Peters, Neumann, Iyyer, Gardner, Clark, Lee, and
  Zettlemoyer}]{peters2018elmo}
Matthew Peters, Mark Neumann, Mohit Iyyer, Matt Gardner, Christopher Clark,
  Kenton Lee, and Luke Zettlemoyer. 2018.
\newblock Deep contextualized word representations.
\newblock In \emph{Proceedings of the 2018 Conference of the North American
  Chapter of the Association for Computational Linguistics: Human Language
  Technologies, Volume 1 (Long Papers)}, volume~1, pages 2227--2237.

\bibitem[{Pylkk{\"a}nen(2008)}]{pylkkanen2008introducing}
Liina Pylkk{\"a}nen. 2008.
\newblock \emph{Introducing arguments}, volume~49.
\newblock MIT press.

\bibitem[{Radford et~al.(2018)Radford, Narasimhan, Salimans, and
  Sutskever}]{radford2018improving}
Alec Radford, Karthik Narasimhan, Tim Salimans, and Ilya Sutskever. 2018.
\newblock Improving language understanding by generative pre-training.
\newblock \emph{URL https://s3-us-west-2. amazonaws.
  com/openai-assets/research-covers/language-unsupervised/language\_
  understanding\_paper. pdf}.

\bibitem[{Sag et~al.(2003)Sag, Wasow, and Bender}]{sag2003syntactic}
Ivan~A Sag, Thomas Wasow, and Emily~M Bender. 2003.
\newblock \emph{Syntactic Theory: A Formal Introduction}, 2 edition.
\newblock CSLI Publications.

\bibitem[{Shi et~al.(2016)Shi, Padhi, and Knight}]{shi2016syntax}
Xing Shi, Inkit Padhi, and Kevin Knight. 2016.
\newblock Does string-based neural {MT} learn source syntax?
\newblock In \emph{Proceedings of the 2016 Conference on Empirical Methods in
  Natural Language Processing}, pages 1526--1534.

\bibitem[{Sportiche et~al.(2013)Sportiche, Koopman, and
  Stabler}]{sportiche2013introduction}
Dominique Sportiche, Hilda Koopman, and Edward Stabler. 2013.
\newblock \emph{An Introduction to Syntactic Analysis and Theory}.
\newblock John Wiley \& Sons.

\bibitem[{Sprouse and Almeida(2012)}]{sprouse2012adger}
Jon Sprouse and Diogo Almeida. 2012.
\newblock Assessing the reliability of textbook data in syntax: Adger's core
  syntax.
\newblock \emph{Journal of Linguistics}, 48(3):609--652.

\bibitem[{Vaswani et~al.(2017)Vaswani, Shazeer, Parmar, Uszkoreit, Jones,
  Gomez, Kaiser, and Polosukhin}]{vaswani2017attention}
Ashish Vaswani, Noam Shazeer, Niki Parmar, Jakob Uszkoreit, Llion Jones,
  Aidan~N Gomez, {\L}ukasz Kaiser, and Illia Polosukhin. 2017.
\newblock Attention is all you need.
\newblock In \emph{Advances in Neural Information Processing Systems}, pages
  5998--6008.

\bibitem[{Wagner et~al.(2009)Wagner, Foster, and van
  Genabith}]{wagner2009grammaticality}
Joachim Wagner, Jennifer Foster, and Josef van Genabith. 2009.
\newblock Judging grammaticality: Experiments in sentence classification.
\newblock \emph{CALICO Journal}, 26(3):474--490.

\bibitem[{Wang et~al.(2018)Wang, Singh, Michael, Hill, Levy, and
  Bowman}]{wang2018glue}
Alex Wang, Amanpreet Singh, Julian Michael, Felix Hill, Omer Levy, and Samuel
  Bowman. 2018.
\newblock Glue: A multi-task benchmark and analysis platform for natural
  language understanding.
\newblock In \emph{Proceedings of the 2018 EMNLP Workshop BlackboxNLP:
  Analyzing and Interpreting Neural Networks for NLP}, pages 353--355.

\bibitem[{Warstadt et~al.(2018)Warstadt, Singh, and
  Bowman}]{warstadt2018neural}
Alex Warstadt, Amanpreet Singh, and Samuel~R Bowman. 2018.
\newblock Neural network acceptability judgments.
\newblock \emph{arXiv preprint arXiv:1805.12471}.

\bibitem[{Wilcox et~al.(2018)Wilcox, Levy, Morita, and Futrell}]{wilcox2018rnn}
Ethan Wilcox, Roger Levy, Takashi Morita, and Richard Futrell. 2018.
\newblock What do rnn language models learn about filler--gap dependencies?
\newblock In \emph{Proceedings of the 2018 EMNLP Workshop BlackboxNLP:
  Analyzing and Interpreting Neural Networks for NLP}, pages 211--221.

\bibitem[{Wilcox et~al.(2019)Wilcox, Qian, Futrell, Ballesteros, and
  Levy}]{wilcox2019structural}
Ethan Wilcox, Peng Qian, Richard Futrell, Miguel Ballesteros, and Roger Levy.
  2019.
\newblock Structural supervision improves learning of non-local grammatical
  dependencies.
\newblock \emph{arXiv preprint arXiv:1903.00943}.

\end{thebibliography}

\newpage
\appendix

\section{Appendix Overview}

This appendix contains detailed descriptions of the 15 major features and 63 minor features appearing in the annotated CoLA analysis set, along with illustrative examples. The full list of features is given in Table 2 of the paper, repeated here as Table \ref{tab:categories}. Examples that are drawn from the analysis set show the line number in parentheses.

\begin{table*}[h!]
    \centering
    \small
    \begin{tabular}{p{0.17\textwidth} p{0.75\textwidth}}
    \toprule
    \textbf{Major Feature ($n$)} & \textbf{Minor Features ($n$)} \\\midrule
    \textbf{Simple (87)} & Simple (87) \\
    \textbf{Pred (256)} & Copula (187),	Pred/SC (45),	Result/Depictive (26)  \\
    \textbf{Adjunct (226)}	&VP Adjunct (162), Misc Adjunct (75), Locative (69), NP Adjunct (52), Temporal (49), Particle (33)\\
    \textbf{Arg Types (428)} & PP Arg VP (242), Oblique (141), PP Arg NP/AP (81),	Expletive (78),	by-Phrase (58)\\
    \textbf{Arg Altern (421)} & High Arity (253),	Passive (114), Drop Arg (112),	Add Arg (91)\\
  \textbf{Imperative (12)} & Imperatives (12) \\
    \textbf{Bind (121)} & Binding:Other (62), Binding:Refl (60) \\
    \textbf{Question (222)} & Emb Q (99),  Pied Piping (80), Rel Clause (76), Matrix Q (56), Island (22) \\
    \textbf{Comp Clause	(190)} & CP Arg VP (110),  No C-izer (41), Deep Embed (30), CP Arg NP/AP (26), Non-finite CP (24), CP Subj (15)\\
    \textbf{Auxiliary (340)}	& Aux (201), Modal (134), Neg (111),   Psuedo-Aux (26)\\
    \textbf{to-VP (170)	}& Control (80), Non-finite VP Misc (38), VP Arg NP/AP	(33), VP+Extract (26), Raising (19) \\
  \textbf{N, Adj (278) }& Compx NP (106), Rel NP (65), Deverbal (53), Trans Adj (39),  NNCompd (35), Rel Adj (26),  Trans NP (21)\\
    \textbf{S-Syntax (286)}& Coord (158), Ellipsis/Anaphor (118), Dislocation (56),  Subordinate/Cond (41), Info Struc (31), S-Adjunct (30), Frag/Paren (9)\\
  \textbf{Determiner	(178)} & Quantifier (139), NPI/FCI (29), Comparative (25), Partitive (18)\\
    \textbf{Violations (145)}& Extra/Missing Expr (65), Infl/Agr Violation (62), Sem Violation (31) \\\bottomrule
    \end{tabular}
    \caption{Major features and their associated minor features (with number of occurrences $n$).}
    \label{tab:categories}
\end{table*}

\section{Simple}
\subsection{Simple}
These are sentences with transitive or intransitive verbs appearing with their default syntax and argument structure. All arguments are noun phrases (DPs), and there are no modifiers, adjuncts, or auxiliaries on DPs or the VP.

\ex. Included
    \a. John owns the book. (37)
    \b. *Us love they. (47)
    \b. The needle poked the cloth. (183)
    
\ex. Excluded
    \a. Cynthia nibbled on the carrot. (200)
    \b. John ate his noodle quietly. (239)
    
\section{Pred (Predicates)}
\subsection{Copulas}
These are sentences including the verb \emph{be} used predicatively. Also, sentences where the object of the verb is itself a predicate, which applies to the subject. Not included are auxiliary uses of \emph{be} or other predicate phrases that are not linked to a subject by a verb.

\ex. Included
\a. John is eager. (27)
\b. He turned into a frog. (150)
\b. To please John is easy. (315)

\ex. Excluded
\a. There is a bench to sit on. (309)
\b. John broke the geode open.
\b. The cake was eaten.

\subsection{Pred/SC (Predicates and Small Clauses)}
These sentences involve predication of a non-subject argument by another non-subject argument, without the presence of a copula. Some of these cases may be analyzed as small clauses.
\cite[see][pp. 189-193]{sportiche2013introduction}

\ex. Included
\a. John called the president a fool. (234)
\b. John considers himself proud of Mary. (464)
\b. They want them arrested. (856)
\b. the election of John president surprised me. (1001)

\subsection{Result/Depictive (Resultatives and Depictives)}
Modifiers that act as predicates of an argument. Resultatives express a resulting state of that argument, and depictives describe that argument during the matrix event. See \cite{goldberg2004resultative}.

\ex. Included
\a. Resultative
\a. *The table was wiped by John clean. (625)
\b. The horse kicked me black and blue. (898)
\z.
\b. Depictive
\a. John left singing. (971)
\b. In which car was the man seen? (398)

\ex. Excluded
\a. He turned into a frog. (150)

\section{Adjunct}

\subsection{Particle}
Particles are lone prepositions associated with verbs. When they appear with transitive verbs they may immediately follow the verb or the object. Verb-particle pairs may have a non-compositional (idiomatic) meaning. See \newcite[pp. 69-70]{carnie2013syntax} and \newcite[pp. 16-17]{kim2008syntax}.

\ex. Included
\a. *The argument was summed by the coach up. (615)
\b. Some sentences go on and on and on. (785)
\b. *He let the cats which were whining out. (71)

\subsection{VP-Adjunct}
Adjuncts modifying verb phrases. Adjuncts are (usually) optional, and they do not change the category of the expression they modify. See \cite[pp.102-106]{sportiche2013introduction}.

\ex. Included
\a. PP-adjuncts, e.g. locative, temporal, instrumental, beneficiary
    \a. Nobody who hates to eat anything should work in a delicatessen. (121)
    \b. Felicia kicked the ball off the bench. (127)
    \z.
\b. Adverbs
    \a. Mary beautifully plays the violin. (40)
    \b. John often meets Mary. (65)
    \z.
\b. Purpose VPs
    \a. We need another run to win. (769)
    \z.

\setlength{\Exlabelwidth}{0.5em}%
\ex. Excluded
\a. PP arguments
    \a. *Sue gave to Bill a book. (42)
    \b. Everything you like is on the table. (736)
    \z.
\b. S-adjuncts
    \a. John lost the race, unfortunately.

\subsection{NP-Adjunct}
These are adjuncts modifying noun phrases. Adjuncts are (usually) optional, and they do not change the category of the expression they modify. Single-word prenominal adjectives are excluded, as are relative clauses (this has another category).
\ex. Included
    \a. PP-adjuncts
        \a. *Tom's dog with one eye attacked Frank's with three legs. (676)
        \b. They were going to meet sometime on Sunday, but the faculty didn't know when. (565)
    \z.
  \b.  Phrasal adjectives
      \a.  As a statesman, scarcely could he do anything worth mentioning. (292)
    \z.
    \b. Verbal modifiers
      \a. The horse raced past the barn fell. (900)

\ex. Excluded
    \a. Prenominal Adjectives
      \a.  It was the policeman met that several young students in the park last night. (227)
    \z.
    \b. Relative Clauses
    \b. NP arguments

\subsection{Temporal}
These are adjuncts of VPs and NPs that specify a time or modify tense or aspect or frequency of an event. Adjuncts are (usually) optional, and they do not change the category of the expression they modify.
\ex. Included
    \a. Short adverbials (never, today, now, always)
        \a. *Which hat did Mike quip that she never wore? (95)
    \z.
    \b. PPs
      \a.  Fiona might be here by 5 o'clock. (426)
    \z.
    \b. When
        \a. I inquired when could we leave. (520)

\subsection{Locative (Locative Adjuncts)}
These are adjuncts of VPs and NPs that specify a location of an event or a part of an event, or of an individual. Adjuncts are (usually) optional, and they do not change the category of the expression they modify.
\ex. Included
    \a. Short adverbials
    \b. PPs
        \a. The bed was slept in. (298)
        \b. *Anson demonized up the Khyber (479)
        \b. Some people consider dogs in my neighborhood dangerous. (802)
        \b. Mary saw the boy walking toward the railroad station. (73)
    \z.
    \b. Where
      \a.  I found the place where we can relax. (307)

\ex. Excluded
    \a. Locative arguments
        \a. *Sam gave the ball out of the basket. (129)
        \b. Jessica loaded boxes on the wagon. (164)
        \b. I went to Rome.

\subsection{Misc Adjunct (Miscellaneous Adjuncts)}
These are adjuncts of VPs and NPs not described by some other category (with the exception of (6-7)), i.e. not temporal, locative, or relative clauses. Adjuncts are (usually) optional, and they do not change the category of the expression they modify.

\ex. Included
    \a. Beneficiary
        \a. *I know which book José didn't read for class, and which book Lilly did it for him. (58)
    \z.
    \b. Instrument
        \a. Lee saw the student with a telescope. (770)
    \z.
    \b. Comitative
        \a. Joan ate dinner with someone but I don't know who. (544)
    \z.
    \b. VP adjuncts
        \a. Which article did Terry file papers without reading? (431)
    \z.
    \b. Purpose
        \a. We need another run to win. (769)

\section{Argument Types}

\subsection{Oblique}
Oblique arguments of verbs are individual-denoting arguments (DPs or PPs) which act as the third argument of verb, i.e. not a subject or (direct) object. They may or may not be marked by a preposition. Obliques are only found in VPs that have three or more individual arguments. Arguments are selected for by the verb, and they are (generally) not optional, though in some cases they may be omitted where they are understood or implicitly existentially quantified over. See \newcite[p.40]{kim2008syntax}.

\ex. Included
    \a. Prepositional
        \a. *Sue gave to Bill a book. (42)
        \b. Mary has always preferred lemons to limes. (70)
        \b. *Janet broke Bill on the finger. (141)
    \z.
    \b. Benefactives
        \a. Martha carved the baby a toy out of wood. (139)
    \z.
    \b. Double object
        \a. Susan told her a story. (875)
    \b. Locative arguments
        \a. Ann may spend her vacation in Italy. (289)
    \z.
    \b. High-arity Passives
        \a. *Mary was given by John the book. (626)

\ex. Excluded
    \a. Non-DP arguments
      \a.  We want John to win (28)
    \z.
    \b. 3rd argments where not all three arguments are DPs
        \a. We want John to win (28)

\subsection{PP Arg VP (PP Arguments of VPs)}
Prepositional Phrase arguments of VPs are individual-denoting arguments of a verb which are marked by a proposition. They may or may not be obliques. Arguments are selected for by the verb, and they are (generally) not optional, though in some cases they may be omitted where they are understood or implicitly existentially quantified over.

\ex. Included
    \a. Dative
        \a. *Sue gave to Bill a book. (42)
    \z.
    \b. Conative (at)
        \a. *Carla slid at the book. (179)
    \z.
    \b. Idiosyncratic prepositional verbs
        \a. I wonder who to place my trust in. (711)
        \b. She voted for herself. (743)
    \z.
    \b. Locative
        \a. John was found in the office. (283)
    \z.
    \b. PP predicates
        \a. Everything you like is on the table. (736)

\ex. Excluded
    \a. PP adjuncts
    \b. Particles
    \b. Arguments of deverbal expressions
        \a. *the putter of books left. (892)
    \z.
    \b. By-phrase
        \a. Ted was bitten by the spider. (613)

\subsection{PP Arg NP/AP (PP Arguments of NPs and APs)}
Prepositional Phrase arguments of NPs or APs are individual-denoting arguments of a noun or adjective which are marked by a proposition. Arguments are selected for by the head, and they are (generally) not optional, though in some cases they may be omitted where they are understood or implicitly existentially quantified over.

\ex. Included
    \a. Relational adjectives
        \a. Many people were fond of Pat. (936)
        \b. *I was already aware of fact. (824)
    \z.
    \b. Relational nouns
        \a. We admired the pictures of us in the album. (759)
        \b. They found the book on the atom. (780)
    \z.
    \b. Arguments of deverbal nouns
        \a. *the putter of books left. (892)

\subsection{By-phrase}
Prepositional arguments introduced with by. Usually, this is the (semantic) subject of a passive verb, but in rare cases it may be the subject of a nominalized verb. Arguments are usually selected for by the head, and they are generally not optional. In this case, the argument introduced with by is semantically selected for by the verb, but it is syntactically optional. See \newcite[p.190]{adger2003core} and \newcite[]{collins2005smuggling}.

\ex. Included
    \a. Passives
        \a. Ted was bitten by the spider. (613)
    \z.
    \b. Subjects of deverbal nouns
        \a. the attempt by John to leave surprised me. (1003)

\subsection{Expletive}
Expletives, or “dummy” arguments, are semantically inert arguments. The most common expletives in English are it and there, although not all occurrences of these items are expletives. Arguments are usually selected for by the head, and they are generally not optional. In this case, the expletive occupies a syntactic argument slot, but it is not semantically selected by the verb, and there is often a syntactic variation without the expletive. See \newcite[p.170-172]{adger2003core} and \newcite[p.82-83]{kim2008syntax}.

\ex. Included
    \a. There---inserted, existential
        \a. *There loved Sandy. (939)
        \b. There is a nurse available. (466)
    \z.
    \b. It---cleft, inserted
        \a. It was a brand new car that he bought. (347)
        \b. It bothers me that John coughs. (314)
        \b. It is nice to go abroad. (47)
    \z.
    \b. Environmental it
        \a. Kerry remarked it was late. (821)
        \b. Poor Bill, it had started to rain and he had no umbrella. (116)
        \b. You've really lived it up. (160)

\ex. Excluded
    \a. John counted on Bill to get there on time. (996)
    \b. I bought it to read. (1026)

\section{Arg Altern (Argument Alternations)}

\subsection{High Arity}
These are verbs with 3 or more arguments of any kind. Arity refers to the number of arguments that a head (or function) selects for. Arguments are usually selected for by the head, and they are generally not optional. They may be DPs, PPs, CPs, VPs, APs or other categories.

\ex. Included
    \a. Ditransitive
        \a. *[Sue] gave [to Bill] [a book]. (42)
        \b. [Martha] carved [the baby] [a toy] out of wood. (139)
    \z.
    \b. VP arguments
        \a. *[We] believed [John] [to be a fountain in the park]. (274)
        \b. [We] made [them] [be rude]. (260)
    \z.
    \b. Particles
        \a. [He] let [the cats which were whining] [out]. (71)
    \z.
    \b. Passives with by-phrase
        \a. *[A good friend] is remained [to me] [by him]. (237)
    \z.
    \b. Expletives
        \a. *[We] expect [there] [to will rain]. (282)
        \b. [There] is [a seat] [available]. (934)
        \b. [It] bothers [me] [that he is here]. (1009)
    \z.
    \b. Small clause
        \a. [John] considers [Bill] [silly]. (1039)

\ex. Excluded
    \a. Results, depictives
        \a. [John] broke [the geode] [open].

\subsection{Drop Arg (Dropped Arguments)}
These are VPs where a canonical argument of the verb is missing. This can be difficult to determine, but in many cases the missing argument is understood with existential quantification or generically, or contextually salient. See \newcite[p.106-109]{sportiche2013introduction}.

\ex. Included
    \a. Middle voice/causative inchoative
        \a. *The problem perceives easily. (66)
    \z.
    \b. Passive
        \a. The car was driven. (296)
    \z.
    \b. Null complement anaphora
        \a. Jean persuaded Robert. (380)
        \b. Nobody told Susan. (883)
    \z.
    \b. Dropped argument
        \a. *Kim put in the box. (253)
        \b. The guests dined. (835)
        \b. I wrote to Bill. (1030)
    \z.
    \b. Transitive adjective
        \a. John is eager. (27)
        \b. We pulled free. (144)
    \z.
    \b. Transitive noun
        \a. I sensed his eagerness. (155)
    \z.
    \b. Expletive insertion
        \a. *It loved Sandy. (949)

\ex. Excluded 
    \a. Ted was bitten by the spider. (613)

\subsection{Add Arg (Added Arguments)}
These are VPs in which a non-canonical argument of the verb has been added. These cases are clearer to identify where the additional argument is a DP. In general, PPs which mark locations, times, beneficiaries, or purposes should be analyzed as adjuncts, while PPs marking causes can be considered arguments. See \newcite[]{pylkkanen2008introducing}.

\ex. Included
    \a. Extra argument
        \a. *Linda winked her lip. (202)
        \b. Sharon fainted from hunger. (204)
        \b. I shaved myself. (526)
    \z.
    \b. Causative
        \a. *I squeaked the door. (207)
    \z.
    \b. Expletive insertion
        \a. There is a monster in Loch Ness. (928)
        \b. It annoys people that dogs bark. (943)
    \z.
    \b. Benefactive
        \a. Martha carved the baby a toy out of wood. (139)

\subsection{Passive}
The passive voice is marked by the demotion of the subject (either complete omission or to a by-phrase) and the verb appearing as a past participle. In the stereotypical construction there is an auxiliary \emph{be} verb, though this may be absent. See \newcite[p.175-190]{kim2008syntax}, \newcite{collins2005smuggling}, and \newcite[p.311-333]{sag2003syntactic}.

\ex. Included
    \a. Verbs
        \a. The earth was believed to be round. (157)
    \z.
    \b. Psuedopassive
        \a. The bed was slept in. (298)
    \z.
    \b. Past participle adjuncts
        \a. The horse raced past the barn fell. (900)

\section{Imperative}

\subsection{Imperative}
The imperative mood is marked by the absence of the a subject and the bare form of the verb, and expresses a command, request, or other directive speech act.

\ex. Included
    \a. *Wash you! (224)
    \b. Somebody just left - guess who. (528)

\section{Binding}

\subsection{Binding:Refl (Binding of Reflexives)}
These are cases in which a reflexive (non-possessive) pronoun, usually bound by an antecedent. See \newcite[p.163-186]{sportiche2013introduction} and \newcite[p.203-226]{sag2003syntactic}.

\ex. Included
    \a. *Ourselves like ourselves. (742)
    \b. Which pictures of himself does John like? (386)

\subsection{Binding:Other (Binding of Other Pronouns)}
These are cases in which a non-reflexive pronoun appears along with its antecedent. This includes donkey anaphora, quantificational binding, and bound possessives, among other bound pronouns. See \newcite[p.163-186]{sportiche2013introduction} and \newcite[p.203-226]{sag2003syntactic}.

\ex. Included
    \a. Bound possessor
        \a. The children admire their mother. (382)
    \z.
    \b. Quantificational binding
        \a. Everybody gets on well with a certain relative, but often only his therapist knows which one. (562)
    \z.
    \b. Bound pronoun
        \a. *We gave us to the cause. (747)

\section{Question}

\subsection{Matrix Q (Matrix Questions)}
These are sentences in which the matrix clause is interrogative (either a wh- or polar question). See \newcite[pp.282-213]{adger2003core}, \newcite[pp.193-222]{kim2008syntax}, and \newcite[p.315-350]{carnie2013syntax}.

\ex. Included
    \a. Wh-question
        \a. Who always drinks milk? (684)
    \z.
    \b. Polar question
      \a.  Did Athena help us? (486)

\subsection{Emb Q (Embedded Questions)}
These are embedded interrogative clauses appearing as arguments of verbs, nouns, and adjectives. Not including relative clauses and free relatives. See \newcite[p.297]{adger2003core}.

\ex. Included
    \a. Under VP
        \a. I forgot how good beer tastes. (235)
        \b. *What did you ask who saw? (508)
    \z.
    \b. Under NP
        \a. That is the reason why he resigned. (313)
    \z.
    \b. Under AP
        \a. They claimed they had settled on something, but it wasn't clear what they had settled on. (529)
    \z.
    \b. Free relative
        \a. What the water did to the bottle was fill it. (33)

\ex. Excluded
    \b. Relative clauses, free relatives

\subsection{Pied Piping}
These are phrasal Wh-phrases, in which the wh-word moves along with other expressions, including prepositions (pied-piping) or nouns in the case of determiner wh-words such as how many and which.

\ex. Included
    \a. Pied-piping
        \a. *The ship sank, but I don't know with what. (541)
    \z.
    \b. Other phrasal wh-phrases
        \a. I know which book Mag read, and which book Bob read my report that you hadn't. (61)
        \b. How sane is Peter? (88)

\subsection{Rel Clause (Relative Clause)}
Relative clauses are noun modifiers appearing with a relativizer (either that or a wh-word) and an associated gap. See \newcite[p.223-244]{kim2008syntax}.

\ex. Included
    \a. Though he may hate those that criticize Carter, it doesn't matter. (332)
    \b. *The book what inspired them was very long. (686)
    \b. Everything you like is on the table. (736)

\ex. Excluded
    \a. *The more you would want, the less you would eat. (6)

\subsection{Island}
This is wh-movement out of an extraction island, or near-island. Islands include, for example, complex NPs, adjuncts, embedded questions, coordination. A near-island is an extraction that closely resembles an island violation, such as extraction out of an embedded clause, or across-the-board extraction. See \newcite[pp.323-333]{adger2003core} and \newcite[pp.332-334]{carnie2013syntax}.

\ex. Included
    \a. Embedded question
        \b. *What did you ask who Medea gave? (493)
    \z.
    \b. Adjunct
      \a.  *What did you leave before they did? (598)
    \z.
    \b. Parasitic gaps
        \a. Which topic did you choose without getting his approval? (311)
    \z.
    \b. Complex NP
        \a. Who did you get an accurate description of? (483)

\section{Comp Clause (Complement Clauses)}

\subsection{CP Subj (CP Subjects)}
These are complement clauses acting as the (syntactic) subject of verbs. See \newcite[pp.90-91]{kim2008syntax}.

\ex. Included
    \a. That dogs bark annoys people. (942)
    \b. The socks are ready for for you to put on to be planned. (112)

\ex. Excluded
    \a. Expletive insertion
        \a. It bothers me that John coughs. (314)

\subsection{CP Arg - VP (CP Arguments of VPs)}
These are complement clauses acting as (non-subject) arguments of verbs. See \newcite[pp.84-90]{kim2008syntax}.

\ex. Included 
    \a. I can't believe Fred won't, either. (50)
    \b. I saw that gas can explode. (222)
    \b. It bothers me that John coughs. (314)
    \b. Clefts
        \a. It was a brand new car that he bought. (347)

\subsection{CP Arg - NP/AP (CP Arguments of NPs and APs)}
These are complement clauses acting as an argument of a noun or adjective. See \newcite[pp.91-94]{kim2008syntax}.

\ex. Included
    \a. Under NP
        \a. Do you believe the claim that somebody was looking for something? (99)
    \z.
    \b. Under AP
        \a. *The children are fond that they have ice cream. (842)

\subsection{Non-Finite CP}
These are complement clauses with a non-finite matrix verb. Often, the complementizer is for, or there is no complementizer. See \newcite[pp.252-253,256-260]{adger2003core}.

\ex. Included
    \a. For complementizer
        \a. I would prefer for John to leave. (990)
    \z.
    \b. No Complementizer
        \a. Mary intended John to go abroad. (48)
    \z.
    \b. Ungrammatical
        \a. Heidi thinks that Andy to eat salmon flavored candy bars. (363)
    \z.
    \b. V-ing
        \a. Only Churchill remembered Churchill giving the Blood, Sweat and Tears speech. (469)

\subsection{No C-izer (No Complementizer)}
These are complement clauses with no overt complementizer.

\ex. Included
    \a. Complement clause
        \a. I'm sure we even got these tickets! (325)
        \b. He announced he would marry the woman he loved most, but none of his relatives could figure out who. (572)
    \z.
    \b. Relative clause
        \a. The Peter we all like was at the party (484)

\subsection{Deep Embed (Deep Embedding)}
These are sentences with three or nested verbs, where VP is not an aux or modal, i.e. with the following syntax: [S \dots [ VP \dots [ VP \dots [ VP \dots  ] \dots ] \dots ] \dots ]

\ex. Included
    \a. Embedded VPs
        \a. Max seemed to be trying to force Ted to leave the room, and Walt, Ira. (657)
    \z.
  \b.  Embedded clauses
        \a. I threw away a book that Sandy thought we had read. (713)

\section{Aux (Auxiliaries)}

\subsection{Neg (Negation)}
Any occurrence of negation in a sentence, including sentential negation, negative quantifiers, and negative adverbs.

\ex. Included
    \a. Sentential
        \a. I can't remember the name of somebody who had misgivings. (123)
    \z.
    \b. Quantifier
        \a. No writer, and no playwright, meets in Vienna. (124)
    \z.
    \b. Adverb
        \a. They realised that never had Sir Thomas been so offended. (409)

\subsection{Modal}
Modal verbs (\emph{may, might, can, could, will, would, shall, should, must}). See \newcite[pp.152-155]{kim2008syntax}.

\ex. Included
    \a. John can kick the ball. (280)
    \b. As a statesman, scarcely could he do anything worth mentioning. (292)

\ex. Excluded
    \a. Pseudo-modals
        \a. Sandy was trying to work out which students would be able to solve a certain problem. (600)

\subsection{Aux (Auxiliaries)}
Auxiliary verbs (e.g. \emph{be, have, do}). See \newcite[pp.149-174]{kim2008syntax}.

\ex. Included
    \a. They love to play golf, but I do not. (290)
    \b. The car was driven. (296)
    \b. he had spent five thousand dollars. (301)

\ex. Excluded
    \a. Pseudo-auxiliaries
      \a.  *Sally asked if somebody was going to fail math class, but I can't remember who. (589)
        \b. The cat got bitten. (926)

\subsection{Psuedo-Aux (Pseudo Auxiliaries)}
These are predicates acting as near-auxiliary (e.g. get-passive) or near-modals (e.g. willing)

\ex. Included
    \a. Near-auxiliaries
        \a. *Mary came to be introduced by the bartender and I also came to be. (55)
        \b. *Sally asked if somebody was going to fail math class, but I can't remember who. (589)
        \b. The cat got bitten. (926)
    \z.
    \b. Near-modals
        \a. Clinton is anxious to find out which budget dilemmas Panetta would be willing to tackle in a certain way, but he won't say in which. (593)
        \b. Sandy was trying to work out which students would be able to solve a certain problem. (600)

\section{to-VP (Infinitival VPs)}

\subsection{Control}
These are VPs with control verbs, where one argument is a non-finite to-VP without a covert subject co-indexed with an argument of the matrix verb. See \newcite[pp.252,266-291]{adger2003core}, \newcite[pp.203-222]{sportiche2013introduction}, and \newcite[pp.125-148]{kim2008syntax}.

\ex. Included
    \a. Intransitive subject control
        \a. *It tries to leave the country. (275)
    \z.
    \b. Transitive subject control
        \a. John promised Bill to leave. (977)
    \z.
    \b. Transitive object control
        \a. I want her to dance. (379)
        \b. John considers Bill to be silly. (1040)

\ex. Excluded 
    \a. VP args of NP/AP
        \a. This violin is difficult to play sonatas on. (114)
    \z.
    \b. Purpose
        \a. There is a bench to sit on. (309)
    \z.
    \b. Subject VPs
        \a. To please John is easy. (315)
    \z.
    \b. Argument present participles
        \a. Medea denied poisoning the phoenix. (490)
    \z.
    \b. Raising
        \a. Anson believed himself to be handsome. (499)

\subsection{Raising}
These are VPs with raising predicates, where one argument is a non-finite to-VP without a covert subject co-indexed with an argument of the matrix verb. Unlike control verbs, the coindexed argument is not a semantic argument of the raising predicate. See \newcite[pp.260-266]{adger2003core}, \newcite[pp.203-222]{sportiche2013introduction}, and \newcite[pp.125-148]{kim2008syntax}.

\ex. Included
    \a. Subject raising
        \a. Under the bed seems to be a fun place to hide. (277)
    \z.
    \b. Object raising
        \a. Anson believed himself to be handsome. (499)
    \z.
    \b. Raising adjective
        \a. John is likely to leave. (370)

\subsection{VP+Extraction (VPs with Extraction)}
These are embedded infinitival VPs containing a (non-subject) gap that is filled by an argument in the upper clause. Examples are purpose-VPs and tough-movement. See \newcite[pp.246-252]{kim2008syntax}.

\ex. Included
    \a. Tough-movement
        \a. *Drowning cats, which is against the law, are hard to rescue. (79)
    \z.
    \b. Infinitival relatives
        \a. *Fed knows which politician her to vote for. (302)
    \z.
    \b. Purpose
         \a. the one with a red cover takes a very long time to read. (352)
    \z.
    \b. Other non-finite VPs with extraction
        \a. As a statesman, scarcely could he do anything worth mentioning. (292)

\subsection{VP arg - NP/AP (VP Arguments of NPs and APs)}
These are non-finite VP arguments of nouns and adjectives.

\ex. Included
    \a. Raising adjectives
        \a. John is likely to leave. (370)
    \z.
    \b. Control adjectives
        \a. The administration has issued a statement that it is willing to meet a student group, but I'm not sure which one. (604)
    \z.
    \b. Control nouns
         \a. As a teacher, you have to deal simultaneously with the administration's pressure on you to succeed, and the children's to be a nice guy. (673)
    \z.
    \b. Purpose VPs
        \a. there is nothing to do. (983)

\subsection{Non-Finite VP Misc (Miscellaneous Infinitival VPs)}
These are miscellaneous non-finite VPs.

\ex. Included
    \a. I saw that gas can explode. (222)
    \b. Gerunds/Present participles
        \a. *Students studying English reads Conrad's Heart of Darkness while at university. (262)
      \b.  Knowing the country well, he took a short cut. (411)
        \b. John became deadly afraid of flying. (440)
    \z.
    \b. Subject VPs
        \a. To please John is easy. (315)
    \z.
    \b. Nominalized VPs
        \a. *What Mary did Bill was give a book. (473)

\ex. Excluded
    \a. to-VPs acting as complements or modifiers of verbs, nouns, or adjectives

\section{N, Adj (Nouns and Adjectives)}

\subsection{Deverbal (Deverbal Nouns and Adjectives)}
These are nouns and adjectives derived from verbs.

\ex. Included
    \a. Deverbal nouns
        \a. *the election of John president surprised me. (1001)
    \z.
    \b. “Light” verbs
        \a. The birds give the worm a tug. (815)
    \z.
    \b. Gerunds
        \a. If only Superman would stop flying planes! (773)
    \z.
    \b. Event-wh
        \a. What the water did to the bottle was fill it. (33)
    \z.
    \b. Deverbal adjectives
        \a. His or her least known work. (95)

\subsection{Rel NP (Relational Nouns)}
Relational nouns are NPs with an obligatory (or existentially closed) argument. A particular relation holds between the members of the extension of NP and the argument. The argument must be a DP possessor or a PP. See \newcite[pp.82-83]{kim2008syntax}.

\ex. Included
    \a. Nouns with of-arguments
        \a. John has a fear of dogs. (353)
    \z.
    \b. Nouns with other PP-arguments
        \a. Henri wants to buy which books about cooking? (442)
    \z.
    \b. Measure nouns
        \a. I bought three quarts of wine and two of Clorox. (667)
    \z.
    \b. Possessed relational nouns
        \a. *John's mother likes himself. (484)

\ex. Excluded
    \a. Nouns with PP modifiers
        \a. Some people consider dogs in my neighborhood dangerous. (802)
        
\subsection{Trans-NP (Transitive NPs)}
Transitive (non-relational) nouns take a VP or CP argument. See \newcite[pp.82-83]{kim2008syntax}.

\ex. Included
    \a. VP argument
        \a. the attempt by John to leave surprised me. (1003)
    \z.
    \b. CP argument
        \a. *Which report that John was incompetent did he submit? (69)
    \z.
    \b. QP argument
         \a. That is the reason why he resigned. (313)

\subsection{Complex NP}
These are complex NPs, including coordinated nouns and nouns with modifiers (excluding prenominal adjectives).

\ex. Included
    \a. Modified NPs
        \a. *The madrigals which Henry plays the lute and sings sound lousy. (84)
        \b. John bought a book on the table. (233)
    \z.
  \b.  NPs with coordination
        \a. *The soundly and furry cat slept. (871)
        \b. The love of my life and mother of my children would never do such a thing. (806)

\subsection{NN Compound (Noun-Noun Compounds)}
Noun-noun compounds are NPs consisting of two constituent nouns.

\ex. Included
    \a. It was the peasant girl who got it. (320)
    \b. A felon was elected to the city council. (938)

\subsection{Rel Adj (Relational Adjectives)}
These are adjectives that take an obligatory (or existentially closed) argument. A particular relation holds between the members of the extension of the modified NP and the argument. The argument must be a DP or PP. See \newcite[pp.80-82]{kim2008syntax}.

\ex. Included
    \a. Of-arguments
        \a. The chickens seem fond of the farmer. (254)
    \z.
    \b. Other PP arguments
        \a. This week will be a difficult one for us. (241)
        \b. John made Bill mad at himself. (1035)

\subsection{Trans- AP (Transitive Adjectives)}
A transitive (non-relational) adjective. I.e. an adjectives that takes a VP or CP argument. See \newcite[pp.80-82]{kim2008syntax}.

\ex. Included
    \a. VP argument
        \a. John is likely to leave. (370)
    \z.
    \b. CP argument
        \a. John is aware of it that Bill is here. (1013)
    \z.
    \b. QP argument
        \a. The administration has issued a statement that it is willing to meet a student group, but I'm not sure which one. (604)

\section{S-Syntax (Sentence-Level Syntax)}

\subsection{Dislocation}
These are expressions with non-canonical word order. See, for example, \newcite[p.76]{sportiche2013introduction}.

\ex. Includes
    \a. Particle shift
        \a. *Mickey looked up it. (24)
    \z.
    \b. Preposed modifiers
        \a. Out of the box jumped a little white rabbit. (215)
        \b. *Because she's so pleasant, as for Mary I really like her. (331)
    \z.
    \b. Quantifier float
        \a. The men will all leave. (43)
    \z.
    \b. Preposed argument
      \a.  With no job would John be happy. (333)
    \z.
    \b. Relative clause extraposition
        \a. Which book's, author did you meet who you liked? (731)
    \z.
    \b. Misplaced phrases
        \a. Mary was given by John the book. (626)

\subsection{Info Struc (Information Structural Movement)}
This includes topicalization and focus constructions. See \newcite[pp.258-269]{kim2008syntax} and \newcite[pp.68-75]{sportiche2013introduction}.

\ex. Included
    \a. Topicalization
        \a. Most elections are quickly forgotten, but the election of 2000, everyone will remember for a long time. (807)
    \z.
    \b. Clefts
        \a. It was a brand new car that he bought. (347)
    \z.
    \b. Pseudo-clefts
        \a. What John promised is to be gentle. (441)

\ex. Excluded
    \a. There-insertion
    \b. Passive

\subsection{Frag/Paren (Fragments and Parentheticals)}
These are parentheticals or fragmentary expressions.
\ex. Included
    \a. Parenthetical
        \a. Mary asked me if, in St. Louis, John could rent a house cheap. (704)
    \z.
    \b. Fragments
        \a. The soup cooks, thickens. (448)
    \z.
    \b. Tag question
        \a. George has spent a lot of money, hasn't he? (291)

\subsection{Coord (Coordination)}
Coordinations and disjunctions are expressions joined with and, but, or, etc. See \newcite[pp.61-68]{sportiche2013introduction}.

\ex. Included
    \a. DP coordination
      \a.  Dave, Dan, Erin, Jaime, and Alina left. (341)
    \z.
    \b. Right Node Raising
        \a. Kim gave a dollar to Bobbie and a dime to Jean. (435)
    \z.
    \b. Clausal coordination
        \a.  She talked to Harry, but I don't know who else. (575)
    \z.
    \b. Or, nor
        \a. *No writer, nor any playwright, meets in Vienna. (125)
    \z.
    \b. Pseudo-coordination
        \a. I want to try and buy some whiskey. (432)
    \z.
    \b. Juxtaposed clauses
        \a. Lights go out at ten. There will be no talking afterwards. (779)

\subsection{Subord/Cond (Subordinate Clauses and Conditionals)}
This includes subordinate clauses, especially with subordinating conjunctions, and conditionals.

\ex. Included
    \a. Conditional
        \a. If I can, I will work on it. (56)
    \z.
    \b. Subordinate clause
        \a. *What did you leave before they did? (598)
        \b. *Because Steve's of a spider's eye had been stolen, I borrowed Fred's diagram of a snake's fang. (677)
    \z.
    \b. Correlative
        \a. *As you eat the most, you want the least. (5)

\subsection{Ellipsis/Anaphora}
This includes VP or NP ellipsis, or anaphora standing for VPs or NPs (not DPs). See \newcite[pp.55-61]{sportiche2013introduction}.

\ex. Included
    \a. VP Ellipsis
        \a. If I can, I will work on it. (56)
        \b. Mary likes to tour art galleries, but Bill hates to. (287)
    \z.
    \b. VP Anaphor
        \a. I saw Bill while you did so Mary. (472)
    \z.
    \b. NP Ellipsis
        \a. Tom's dog with one eye attacked Fred's. (679)
    \z.
    \b. NP anaphor
        \a. the one with a red cover takes a very long time to read. (352)
    \z.
    \b. Sluicing
        \a. Most columnists claim that a senior White House official has been briefing them, and the newspaper today reveals which one. (557)
    \z.
    \b. Gapping
        \a. Bill ate the peaches, but Harry the grapes. (646)

\subsection{S-adjunct (Sentence-Level Adjuncts)}
These are adjuncts modifying sentences, sentence-level adverbs, subordinate clauses.

\ex. Included
    \a. Sentence-level adverbs
        \a. Suddenly, there arrived two inspectors from the INS. (447)
    \z.
    \b. Subordinate clauses
        \a. The storm arrived while we ate lunch. (852)

\section{Determiner}

\subsection{Quantifier}
These are quantificational DPs, i.e. the determiner is a quantifier.

\ex. Included
    \a. Quantifiers
        \a. *Every student, and he wears socks, is a swinger. (118)
        \b. We need another run to win. (769)
    \z.
    \b. Partitive
        \a. *Neither of students failed. (265)

\subsection{Partitive}
These are quantifiers that take PP arguments, and measure nouns. See \newcite[pp.109-118]{kim2008syntax}.

\ex. Included
    \a. Quantifiers with PP arguments
        \a. *Neither of students failed. (265)
    \z.
    \b. Numerals
        \a. One of Korea's most famous poets wrote these lines. (294)
    \z.
  \b.  Measure nouns
        \a. I bought three quarts of wine and two of Clorox. (667)

\subsection{NPI/FCI (Negative Polarity and Free Choice Items)}
These are negative polarity items (any, ever, etc.) and free choice items (any). See \newcite{kadmon1993any}.

\ex. Included
    \a. NPI
        \a. Everybody around here who ever buys anything on credit talks in his sleep. (122)
        \b. I didn't have a red cent. (350)
    \z.
    \b. FCI
        \a. Any owl hunts mice. (387)

\subsection{Comparative}
These are comparative constructions. See \cite{culicover1999comparative}.

\ex. Included
    \a. Correlative
        \a. The angrier Mary got, the more she looked at pictures. (9)
        \b. They may grow as high as bamboo. (337)
        \b. I know you like the back of my hand. (775)

\section{Violations}

\subsection{Sem Violation (Semantic Violations)}
These are sentences that include a semantic violation, including type mismatches, violations of selectional restrictions, polarity violations, definiteness violations. 

\ex. Included
    \a. Volation of selectional restrictions
        \a. *many information was provided. (218)
        \b. *It tries to leave the country. (275)
    \z.
    \b. Aspectual violations
        \a. *John is tall on several occasions. (540)
    \z.
    \b. Definiteness violations
        \a. *It is the problem that he is here. (1018)
    \z.
    \b. Polarity violations
        \a. Any man didn't eat dinner. (388)

\subsection{Infl/Agr violation (Inflection and Agreement Violations)}
These are sentences that include a violation in inflectional morphology, including tense-aspect marking, or agreement.

\ex. Included
    \a. Case
        \a. *Us love they. (46)
    \z.
    \b. Agreement
        \a. *Students studying English reads Conrad's Heart of Darkness while at university. (262)
    \z.
    \b. Gender
        \a. *Sally kissed himself. (339)
    \z.
    \b. Tense/Aspect
        \a. *Kim alienated cats and beating his dog. (429)

\subsection{Extra/Missing Word}
These are sentences with a violation that can be identified with the presence or absence of a single word.

\ex. Included
    \a. Missing word
        \a. *John put under the bathtub. (247)
        \b. *I noticed the. (788)
    \z.
    \b. Extra word
        \a. *Everyone hopes everyone to sleep. (467)
        \b. *He can will go (510)

\end{document}